\documentclass{article} 
\usepackage{morpheus2,times}

\usepackage{amsmath,amsfonts,bm}

\def\eqref#1{equation~\ref{#1}}

\def\1{\bm{1}}

\DeclareMathAlphabet{\mathsfit}{\encodingdefault}{\sfdefault}{m}{sl}
\SetMathAlphabet{\mathsfit}{bold}{\encodingdefault}{\sfdefault}{bx}{n}

\usepackage{hyperref}
\usepackage{url}
\usepackage{amsmath,amssymb}
\usepackage{booktabs}
\usepackage{graphicx}
\usepackage{xcolor}
\usepackage{caption}

\newcommand{\morpheustwo}{\textsc{Morpheus-2.0}}
\newcommand{\morpheusmultiscale}{\textsc{Morpheus-2.0 Multiscale}}
\newcommand{\appref}[2]{Appendix~\ref{app:#1}}
\newcommand{\mainref}[2]{Section~\ref{#1} of the main paper}

\finalcopy 

\title{\morpheustwo{}: Multiscale Probabilistic Forecasting with Dual-Axis Attention}

\author{
Aakriti\thanks{Equal contribution. Corresponding author: Aakriti
(\texttt{aakriti.aakriti@jpmorgan.com}).} \And
Manoj Cherukumalli\footnotemark[1] \And
Leonidas Tsepenekas \And
Prathamesh Patil \AND
Penghang Liu \And
Elizabeth Fons \And
Amit Varshney \And
Vamsi Potluru
}

\begin{document}
\maketitle

\makeatletter
\lhead{\morpheustwo{}: Multiscale Probabilistic Forecasting with Dual-Axis Attention}
\makeatother

\begin{abstract}
A universal forecaster is trained once and applied everywhere, so it must serve
tasks whose schemas, forecast horizons, and temporal resolutions all differ. The
number of targets and covariates changes from dataset to dataset, horizons vary
by orders of magnitude, and sampling rates range from seconds to years. Existing
time-series foundation models address these only in part, since they isolate
each series or visit the two grid relations in separate blocks, and they fix the
horizon and resolution before training begins. We introduce \textbf{\morpheustwo{}}, a compact
encoder-only foundation model that represents targets and covariates on a single
variable--time grid. Its dual-axis attention runs both relations side by side in
every layer, one across variables at a timestep and one across time within a
variable; a learned gate weights the two per layer, and an
axis-aware rotary encoding preserves both grid coordinates. The forecast length
is sampled during training, so one set of weights spans short
to long horizons at no additional inference cost. \textbf{\morpheusmultiscale{}}
extends the model to several temporal resolutions at once, weighting them
according to the series being forecast. Across GIFT-Eval, which spans
several domains and frequencies, the two releases give complementary operating
points: \morpheustwo{} attains $0.685$ relative MASE with 81M
parameters, while the larger \morpheusmultiscale{} attains
$\mathbf{0.646}$ relative MASE and $\mathbf{0.455}$ relative CRPS, the lowest
of any published system on both metrics.
\end{abstract}

\section{Introduction}
Forecasting supports decisions in energy, retail, finance, climate, and cloud
operations, yet conventional systems are tuned to a particular dataset,
frequency, and horizon, which makes broad deployment expensive to maintain.
Time-series foundation models (TSFMs) offer a different paradigm: pretraining on
heterogeneous series lets one model transfer across domains with little or no
task-specific optimization~\citep{chronos,timesfm,moirai,toto}. The central
problem then becomes building a single model whose representation and objective
remain effective across that heterogeneity.

That heterogeneity is structural, and it runs along three axes. A forecasting
example forms a two-dimensional variable--time grid, and the first axis is its
\emph{schema}: how many rows that grid has, its \emph{schema width}, and what
each row is, whether a target to be predicted, a past-only covariate, or a
covariate known into the future~\citep{chronos2,toto2,tirex2}. Both change from
task to task. Channel-independent processing
sidesteps this but cannot exploit cross-variable structure, while unrestricted
attention over the grid pays quadratic cost on both axes and provides no
inductive bias to separate relations
across variables and dynamics within a variable.

The second and third axes are both temporal, the forecast horizon and the patch resolution. One is the forecast
horizon: training at a single fixed length over-specializes the model, since
short output blocks force iterative rollout by feeding forecast back into context,
while very wide blocks make every training prediction harder. The other is temporal
resolution. We group each variable into consecutive blocks of $s$ timesteps, a
\emph{patch}, and treat each as one token; $s$ is the \emph{patch width}.
Narrow patches resolve brief events, wide ones reach further back and smooth
what they cover, so no single choice suits data sampled from seconds to years.
Section~\ref{sec:analysis} examines both effects.

We introduce \textbf{\morpheustwo{}}, an encoder-only TSFM that casts
forecasting as masked prediction over an interleaved variable--time grid, with
covariates and targets sharing one flexible-schema representation and masks
enforcing the information available at the forecast origin. In every layer,
dual-axis attention runs two pathways, one across variables at the same
timestep and one across time within the same variable, fused by a learned gate.
This yields $O(TN^2+T^2N)$ permitted pairs for $T$ times and $N$
variables instead of $O(T^2N^2)$, executed with block-sparse
FlexAttention~\citep{flexattention}, while axis-aware two-dimensional RoPE
preserves the grid coordinates. Prediction-length randomization samples
the forecast length for each training batch, so one parameter set covers short
and long requests alike, with no horizon-specific heads and no added inference
cost.

We further introduce \textbf{\morpheusmultiscale{}}, which combines several
\morpheustwo{} encoders trained at different patch widths and with different
hyperparameters, maps the observed context and requested horizon to a weighting
over them, and pools their marginal forecasts. Across the 97 configurations of
GIFT-Eval~\citep{gifteval}, the 81M-parameter \morpheustwo{} reaches $0.685$
relative mean absolute scaled error (MASE), competitive with substantially
larger systems, and \morpheusmultiscale{} reaches $0.646$ relative MASE and
$0.455$ relative CRPS, the lowest of any publicly reported system on both
(Table~\ref{tab:main}).

Our main contributions are:
\begin{list}{$\bullet$}{
  \setlength{\leftmargin}{1.1em}
  \setlength{\itemindent}{0pt}
  \setlength{\labelsep}{0.45em}
  \setlength{\itemsep}{0.15em}
  \setlength{\parsep}{0pt}
  \setlength{\partopsep}{0pt}
  \setlength{\topsep}{0.25em}}
  \item A novel two-dimensional variable--time representation with
        axis-factorized sparse attention, gated cross-axis fusion, and
        axis-aware two-dimensional RoPE.
  \item Prediction-length randomization for training one probabilistic model
        across forecast horizons without additional inference cost.
  \item Two released models at complementary operating points: a compact
        \morpheustwo{}, and \morpheusmultiscale{}, which weights encoders of
        several patch widths per instance to reach the lowest published
        relative MASE and CRPS on GIFT-Eval.
\end{list}
\vspace{-0.9em}

\section{Related Work}\label{sec:related}
Statistical baselines under automatic order selection remain
competitive~\citep{hyndman2008}, and the M4 competition was won by a hybrid of
exponential smoothing and a recurrent network~\citep{m4}. Deep models displaced
them by training one network across many series~\citep{deepar,nbeats}, with the
Temporal Fusion Transformer adding covariate pathways~\citep{tft}, until
LTSF-Linear matched the long-horizon transformers that followed with a one-layer
linear map~\citep{dlinear}.

Foundation models remove the per-dataset fitting that the models above require: one
set of pretrained weights forecasts unseen series, which leaves open how to
represent a series generally enough for that to work. Chronos
discretizes scaled values into token sequences, TimesFM uses a patched decoder
with direct output patches, and Moirai broadened this to masked, any-variate
probabilistic forecasting~\citep{chronos,timesfm,moirai}; TiRex instead adopts
an xLSTM backbone and Sundial replaces discrete tokenization with flow
matching~\citep{tirex,sundial}. Recent systems model variables jointly, with
Chronos-2 and TiRex-2 supporting multivariate targets alongside past-only and
known-future covariates and Toto 2.0 keeping a multivariate time/variate
backbone~\citep{chronos2,tirex2,toto,toto2}. \morpheustwo{} follows
Morpheus~\citep{morpheus1} and these systems in adopting the covariate-aware
multivariate setting.

How that setting is represented varies widely. Channel-independent patching
treats each series on its own, robust but blind to contemporaneous
signal~\citep{patchtst}; iTransformer makes a whole variate one token, and
Crossformer keeps both axes but visits them in
stages~\citep{itransformer,crossformer}, instances of the axial
factorization template~\citep{axialattention}. Among foundation models, Moirai
and Morpheus flatten variable--time patches into one sequence, Chronos-2 patches
series separately and attends within a related group, and TiRex-2 keeps a
variate-by-patch tensor through alternating
mixers~\citep{moirai,morpheus1,chronos2,tirex2}. Several visit the two axes in
turn; \morpheustwo{} evaluates both in every layer and learns their balance at
each depth.

A second axis of variation is how the horizon and the temporal scale are
supervised. MQ-RNN established direct multi-horizon quantile
heads~\citep{mqrnn}, and TimesFM moved from autoregressive output patches to
filling a masked horizon in one pass~\citep{timesfm,timesfm3}. Training also
varies the span, through context--prediction splits, output patch counts, or
variable-length masked spans~\citep{moirai,chronos2,toto2}, and across scales
Moirai learns patch-size-specific projections, Tiny Time Mixers pairs adaptive
patching with resolution sampling, and N-HiTS uses multi-rate sampling with
hierarchical interpolation~\citep{moirai,ttm,nhits}. Data is the other limiting
resource: public corpora have grown largely by restacking the pool Monash
standardized~\citep{monash,moirai,timer,timemoe,sundial}, so the independent
sources beneath them are fewer than their totals suggest, and generated series
are now a standard component of pretraining mixtures~\citep{chronos,timesfm}.
\morpheustwo{} follows this practice with MySTiC
(Section~\ref{sec:setup-data}).

\section{Problem Formulation}\label{sec:problem}
A forecast is issued at time $T$ for the next $H$ timesteps. Each example
contains $N$ variables on a common time axis, and each variable is one of three
types: $N_y\geq1$ \emph{targets}, whose future values are predicted; $N_p$
\emph{past-only covariates}, observed up to $T$; and $N_f$ \emph{known-future
covariates}, such as calendar features or planned controls, available up to
$T+H$. Every value carries a binary observation mask that separates missing
entries from observed zeros. We write $\mathcal C_T$ for everything the model
may use when the forecast is issued: the history of every variable up to $T$ and
the known-future covariates up to $T+H$. The schema $(N_y,N_p,N_f)$, the context
length, the horizon, and the missingness pattern all vary across examples, and
we refer to each variable as a \emph{row} of the grid representation of
Section~\ref{sec:method-representation}.

For each target $j$ and lead time $h$, a single shared model predicts quantiles
of $y_{j,T+h}$ given $\mathcal C_T$: the $\tau$-quantile is the value that
$y_{j,T+h}$ falls at or below with probability $\tau$, and the model predicts it
as $\widehat Q_{j,h}(\tau)$ for every level $\tau$ in a fixed grid $\mathcal Q$
(\appref{quantile-evaluation}{C3}). The median $\widehat Q_{j,h}(0.5)$ is the
point forecast, and the remaining levels describe uncertainty around it.
Formally, we seek a single parameter vector $\theta$ whose map
$f_\theta:(\mathcal C_T,H)\mapsto\{\widehat Q_{j,h}(\tau)\}$ minimizes the
expected quantile loss (Section~\ref{sec:method-masked}) across schemas,
horizons, and sampling frequencies.

\section{Methodology}\label{sec:method}

\begin{figure}[t]
      \centering
      \includegraphics[width=\textwidth]{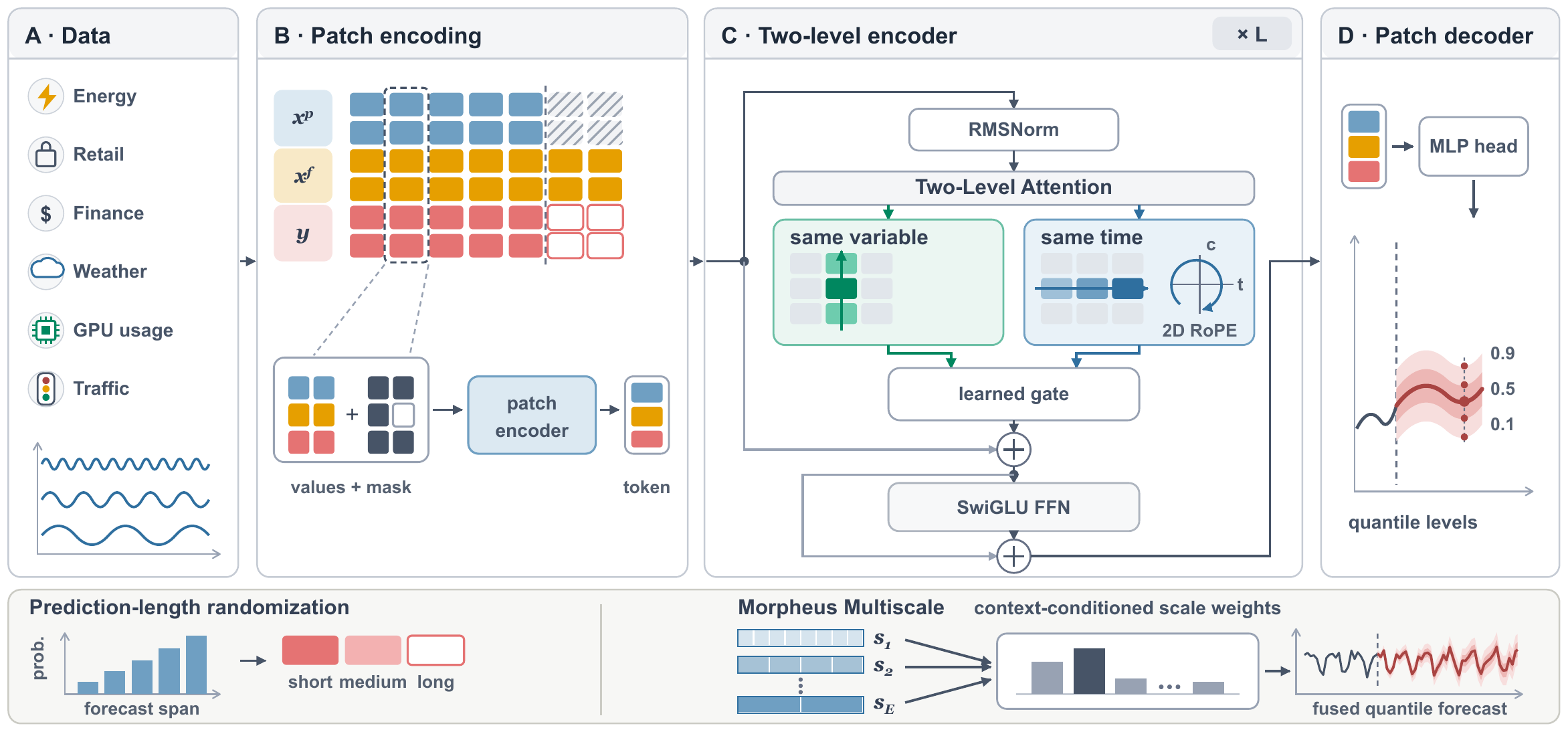}
      \caption{Overview of \morpheustwo{}. \textbf{(A)} Training data span many
      domains, frequencies, and schemas. \textbf{(B)} Covariates and targets form
      one variable--time grid; each patch is embedded with its observation mask,
      so one time patch across all rows becomes one token, and future targets
      remain query-only. \textbf{(C)} Every block applies two parallel attention
      pathways, same-time across variables and same-variable across time with
      axis-aware two-dimensional RoPE, combined by a learned scalar gate.
      \textbf{(D)} The patch decoder projects each token to the native quantile
      grid, with one prediction length sampled per batch. \morpheusmultiscale{}
      runs several such encoders at different patch widths and pools their
      forecasts (Section~\ref{sec:method-multiscale}).}
      \label{fig:architecture}
\end{figure}

Figure~\ref{fig:architecture} summarizes the architecture. \morpheustwo{} is an
encoder-only transformer trained under one pinball
objective~\citep{koenker1978} and following the masked-encoder formulation of
Moirai and Chronos-2~\citep{moirai,chronos2}.

\subsection{Flexible-Schema Variable--Time Representation}\label{sec:method-representation}

Let $s$ denote the patch width and $n\in\{1,\ldots,N\}$ index the concatenated
past-covariate, known-future-covariate, and target rows. For temporal patch $k$,
\morpheustwo{} embeds the value vector $\mathbf{v}_{k,n}\in\mathbb{R}^{s}$
concatenated with the matching slice $\mathbf{m}_{k,n}\in\{0,1\}^{s}$ of the
observation masks of Section~\ref{sec:problem}, giving one token
$\mathbf{z}_{k,n}=\operatorname{PatchEmbed}([\mathbf{v}_{k,n};\mathbf{m}_{k,n}])$.
Tokens are flattened in time-major order, so all variables at a temporal patch
are contiguous, while explicit time, variable, and sample identifiers retain the
grid structure, and shared patch and output projections permit $N$ to vary
without dataset-specific heads. A fixed token budget $P$ bounds this
flexibility: at most $\lfloor P/N\rfloor$ temporal patches are available per
row. Most practical tasks have $N\le50$, and the model accepts any $N$, with
fewer temporal patches per row as $N$ grows.

\subsection{Masked-Horizon Quantile Forecasting}\label{sec:method-masked}

\morpheustwo{} places context and forecast positions in one encoder sequence. A
token $j$ is an admissible key exactly when $n_j$ is a known-future row or
$t_j\le T$: history is visible bidirectionally, future targets and past-only
covariates become masked placeholders that remain queries, and known-future
covariates stay visible past $T$ because the task supplies them. Both
neighborhoods of Section~\ref{sec:method-tla} are intersected with that key set
and the loss is evaluated only at future target positions, so each target query
attends to everything available at the forecast origin without reaching another
target's future value along either axis. At each masked target position the
output head directly predicts marginal quantiles
$\{\widehat Q_{j,h}(\tau):\tau\in\mathcal Q\}$, trained with the pinball loss
$\rho_\tau(u)=\max\{\tau u,(\tau-1)u\}$ on
$u=y-\widehat Q(\tau)$~\citep{koenker1978}, averaged over observed future
targets and summed over levels. The whole native block is emitted in one forward
pass, and longer requests append the predicted medians to the context and
repeat; this direct multi-horizon quantile
interface~\citep{mqrnn,chronos2,toto2} is what lets different patch widths be
aligned and combined.

\subsection{Dual-Axis Attention}\label{sec:method-tla}

Let token $i$ have temporal index $t_i$ and variable index $n_i$, the row index
of Section~\ref{sec:method-representation}. Its intra-time and inter-time
neighborhoods are $\mathcal I(i)=\{j:t_j=t_i\}$ and
$\mathcal T(i)=\{j:n_j=n_i\}$, with padding and cross-sample pairs removed.
Independent projections attend over each, giving
$\mathbf{a}^{\mathcal I}_i$ and $\mathbf{a}^{\mathcal T}_i$, which layer $\ell$
fuses through output projections $W^{\mathcal I}$ and $W^{\mathcal T}$ as
\begin{equation}
\mathbf{u}_i=(1-g_\ell)W^{\mathcal I}\mathbf{a}^{\mathcal I}_i
              +g_\ell W^{\mathcal T}\mathbf{a}^{\mathcal T}_i,
\qquad g_\ell=\operatorname{sigmoid}(\gamma_\ell),
\end{equation}
with $\gamma_\ell$ one unconstrained trainable
scalar per layer, learned from the forecasting loss alone
(\appref{architecture}{A1}). A fixed $g_\ell=1/2$ would hold one balance at
every depth; nine learned scalars let each block set its own balance
(Section~\ref{sec:ablation-tla}).

Evaluating both axes inside one layer is the property we import from vision
transformers, whose patch tokens keep their row--column coordinates so any layer
can relate two patches along either axis~\citep{vit}. A forecasting example has
the same structure, so we keep simultaneous access to both axes while replacing
attention over all $(TN)^2$ pairs with attention over
$\mathcal I(i)\cup\mathcal T(i)$. Recent forecasting backbones instead alternate
the two relations across
blocks~\citep{axialattention,crossformer,chronos2,toto2,tirex2}, updating a
token by one relation at a time, so a dependence coupling the two axes is never
formed inside a block and survives only if the intermediate representation
carries the first relation forward to the block supplying the second. This
matters most for covariate-conditioned forecasting, where a target must
associate a covariate observed now with the temporal pattern carried by that
covariate's history; Section~\ref{sec:ablation-tla} isolates the operator at
matched capacity. Restricting attention to pairs that share a coordinate also
reduces cost, since the two neighborhoods admit $TN^2+NT^2$ pairs compared with
$(TN)^2$ under unrestricted attention, and the difference grows with the size of
the grid. Block-sparse FlexAttention realizes the retained pairs directly instead of
masking a dense score matrix, so the discarded ones are never materialized.

\subsection{Positional Encoding}\label{sec:method-position}

Attention ignores token order, so each token must carry its grid position.
RoPE~\citep{roformer,ropevision} makes the attention score depend only on the
offset between two tokens, but applied to the flattened index $p=tN+n$ it ties
time offsets to the schema width $N$, and applied to time alone it tells the
intra-time pathway nothing about which variables it compares. We rotate the
$r$-th rotary pair of each query and key by
\begin{equation}
\phi_r(t,n)=t\,\omega_r+\alpha\,n\,\omega_r,
\end{equation}
where $t$ is the time index, $n$ the variable index, $\omega_r$ the standard
RoPE frequencies, and $\alpha=0.04$. The intra-time pathway compares tokens with
equal $t$ and so sees only $\Delta n$; the inter-time pathway compares tokens
with equal $n$ and so sees only $\Delta t$. The two coordinates therefore cannot
alias (\appref{position-impl}{A2}). $\alpha$ and the rotary base are fixed by a
sweep on held-out synthetic data; the inter-time pathway also adds a learned relative-position bias over
time.

\subsection{Prediction-Length Randomization}\label{sec:method-ps}

Requested horizons run from minutes of grid load to quarters of financial
planning. Training
at one fixed prediction length optimizes exactly that span at every step:
shorter requests are never supervised on their own, and longer ones must be
reached by iterative rollout, which accumulates error. Narrowing the
block to fix the first problem only worsens the second.

Each training batch samples one prediction-token budget $M$, the number of
prediction patches, and derives a per-example horizon from it: for an example
with $N_i$ interleaved rows and patch width $s$,
\begin{equation}
M\in\{m_1,\ldots,m_K\},\quad \Pr(M=m_k)=\pi_k,\qquad
H_i(M)=s\max\!\left(1,\left\lfloor M/N_i\right\rfloor\right),
\label{eq:schema-horizon}
\end{equation}
and training optimizes $\mathbb{E}_{M}\mathbb{E}_{i}[\mathcal L_{H_i(M)}]$. A
shared budget keeps the packed layout uniform within the batch, and because the
$M$ prediction tokens are spread over $N_i$ rows, one draw yields different
horizons for examples of different widths, so a single parameter set is trained
across both horizons and schema widths. Related models vary
context--prediction boundaries, output-patch counts, or contiguous masked
spans~\citep{moirai,chronos2,toto2,tirex2}; contiguous masking hides a span
inside the series and recovers it from values on both sides, whereas we always
mask the final span, so every sampled task is a forecast. At inference a
requested horizon is served by masking that many future positions.

\subsection{\morpheusmultiscale{}: Context-Conditioned Scale Fusion}
\label{sec:method-multiscale}

A \morpheustwo{} encoder with patch width $s_e$ produces marginal quantiles
$\widehat Q_e(\tau\mid\mathcal C_T)$ for every target and lead time, so encoders
trained at different widths emit aligned outputs that can be combined directly.
\morpheusmultiscale{} combines $E\geq2$ such encoders, trained at different
patch widths and with different hyperparameters: narrow widths resolve local
detail, wide widths longer temporal structure. For each forecast, a
\emph{mixing head} maps the context $\mathcal C_T$ and the requested horizon $H$
to nonnegative weights $w_e(\mathcal C_T,H)$ that sum to one, and the forecast
is the linear pool of the encoders' predictive distributions,
\begin{equation}
\widehat F_{\mathrm{multiscale}}(x\mid\mathcal C_T,H)
=\sum_{e=1}^{E}w_e(\mathcal C_T,H)\,\widehat F_e(x\mid\mathcal C_T),
\end{equation}
where $\widehat F_e$ interpolates encoder $e$'s quantiles linearly between
levels and reported quantiles are read off by inverting the pool
(\appref{scale-training}{C2}). The same weights apply to every target and lead
time. Pooling distributions, unlike averaging quantile
functions~\citep{lichtendahl2013}, widens the forecast where the encoders
disagree, and since CRPS is convex in the predictive distribution, the pool
never scores worse than the weighted average of its
encoders~\citep{gneiting2007}.

The mixing head is a dilated temporal convolutional network over each target's
normalized context, pooled over time and then over targets so that it accepts
any schema, and combined with an embedding of the requested horizon. The
encoders are trained first; the mixing head is then trained over them by
cross-entropy toward the encoders with the lowest quantile loss on each training
window (\appref{scale-training}{C2}).

\section{Experimental Setup}\label{sec:setup}

\subsection{Pretraining Data}\label{sec:setup-data}
\morpheustwo{} is pretrained on a mixture of real and synthetic series. The real
portion is the public GIFT-Eval pretraining corpus, roughly 230 billion
observations across diverse domains and frequencies~\citep{gifteval}; we train
on all of it, assigning each domain--frequency combination its own sampling
weight, since a few very large sources would otherwise drown out the rest.

Real collections alone leave several regimes underrepresented, in particular
wide multivariate schemas, informative known-future covariates, and controlled
degrees of nonstationarity. We therefore construct the \textbf{Multivariate
Synthetic Time-series Collection (MySTiC)}, a generated corpus supplying those
regimes through three subsets, the first of which extends the generator of
Morpheus~\citep{morpheus1}: composed Gaussian-process kernels with factor and
lead--lag coupling, giving correlated panels at every schema width; covariates
attached to targets through known response functions, so covariate-conditioned
supervision has its dependence fixed by construction; and emulated signals that
reproduce varied temporal structure (\appref{mystic}{B2}).

\subsection{Training Details}\label{sec:setup-training}
All reported models instantiate one encoder architecture and differ in patch
width, prediction-token budget, and training hyperparameters
(\appref{model-manifests}{A3}). We train
each with AdamW~\citep{adamw} and a one-cycle learning-rate
schedule~\citep{smith2018onecycle}, using standard multi-GPU data parallelism
and bfloat16 arithmetic, with training budgets set by realized patch-token
exposure per trainable parameter~\citep{hoffmann2022training}. Hyperparameters
are tuned on held-out synthetic multivariate data.

The four-stage data curriculum progressively increases distributional complexity
and realism, moving from elementary synthetic dynamics through structurally
richer synthetic processes to a final stage dominated by real observations, so
that reusable temporal primitives are acquired before the model must contend
with the noise and irregularity of observed series. \morpheusmultiscale{} adds
one further stage, in which the mixing head of
Section~\ref{sec:method-multiscale} is fitted on the GIFT-Eval training split on
top of the trained encoders.
Stage composition appears in \appref{data-schedule}{B3}, and packing,
optimization, and hardware in \appref{optimization}{A4} and
\appref{model-manifests}{A3}.

\subsection{Benchmark, Metrics, and Protocol}\label{sec:setup-benchmarks}\label{sec:setup-metrics}
GIFT-Eval~\citep{gifteval} contains 23 datasets, more than 144,000 series, and
177 million observations across seven domains, with ten frequencies spanning
seconds to years, univariate and multivariate data, and multiple horizons,
giving 97 configurations with seasonal, trending, intermittent, bursty,
regime-changing, and cross-variable dynamics. We score every test item under
GIFT-Eval's rolling-origin protocol, as do all published results we compare
against.

We report two headline metrics, with MSE, MAPE, sMAPE, and ND in
\appref{additional-metrics}{E2}. Point accuracy is the mean absolute scaled
error (MASE): the absolute error of the median forecast divided by the in-sample
error of a seasonal naive predictor on the same item, which makes it scale free
across series of very different magnitude~\citep{hyndman2006mase}.
Probabilistic accuracy is the continuous ranked probability score (CRPS), which
GIFT-Eval approximates by the mean weighted quantile loss over the 9 levels
$\{0.1,\ldots,0.9\}$, normalized by total absolute label
mass~\citep{gifteval,gluonts}.
Both metrics are aggregated within each configuration, then reported as a
geometric mean (GM) over the 97 configurations relative to Seasonal Naive, which
therefore scores $1.0$; longer requests use iterative
rollout (\appref{quantile-evaluation}{C3}).

\section{Main Results}\label{sec:results}

\begin{table}[t]
\centering
\small
\caption{GIFT-Eval results over 97 configurations. Rel.\ MASE scores the median
forecast and Rel.\ CRPS the full quantile forecast, both GM
relative to Seasonal Naive; lower is better. Setup records whether benchmark
training data are used; parameter counts are taken from the official benchmark,
and ours from the released checkpoints.}
\label{tab:main}
\setlength{\tabcolsep}{7.0pt}
\begin{tabular}{l r l r r}
\toprule
Model & Params & Setup & Rel. MASE $\downarrow$ & Rel. CRPS $\downarrow$ \\
\midrule
Seasonal Naive          & --       & Reference & 1.000 & 1.000 \\
Moirai-2.0-R-Small      & 11.4M    & Pretrained & 0.728 & 0.516 \\
TimesFM-2.5             & 200M     & Zero-shot & 0.705 & 0.490 \\
Chronos-2               & 120M     & Pretrained & 0.698 & 0.485 \\
\textbf{\morpheustwo{}} & 81.5M & Pretrained & 0.685 & 0.519 \\
Toto-2.0-2.5B-FT       & 2.5B     & Fine-tuned & 0.679 & 0.463 \\
TiRex-2-Pretrained      & 82.5M    & Pretrained & 0.678 & 0.467 \\
Toto-2.0-FnF            & N/A      & Pretrained & 0.676 & 0.463 \\
TimesFM-3               & 330M     & Zero-shot & 0.667 & \underline{0.456} \\
Falcon-2.0              & 585M     & Pretrained & \underline{0.666} & 0.486 \\
\textbf{\morpheusmultiscale{}} & 413M & Pretrained & \textbf{0.646} & \textbf{0.455} \\
\bottomrule
\end{tabular}
\vspace{-0.8em}
\end{table}

Table~\ref{tab:main} compares both releases with public GIFT-Eval submissions
scored under the same protocol; the full leaderboard and a per-domain breakdown
are in \appref{gifteval-extended}{E1} and \appref{domain-breakdown}{E3}.
\morpheusmultiscale{} combines five \morpheustwo{} encoders of 80--84M
parameters each, trained at patch widths 2, 4, and 8 with different
hyperparameters, under a 1.4M-parameter mixing head
(\appref{model-manifests}{A3}). It reaches the lowest relative MASE and the
lowest relative CRPS of any published system.

Both releases outrank larger systems trained under the same setup:
\morpheusmultiscale{} improves on Falcon-2.0 by $3.0\%$ relative MASE with 71\%
of its parameters, and \morpheustwo{} is more accurate than Chronos-2 with a
third fewer parameters. Combining encoders improves on the standalone model by
$5.8\%$ in MASE and $12.4\%$ in CRPS, in every horizon group. Against
TimesFM-3, the strongest prior system on five of the six metrics of
\appref{additional-metrics}{E2}, \morpheusmultiscale{} is $3.1\%$ ahead
in MASE and level in CRPS. By horizon, \morpheusmultiscale{} is the
most accurate system on medium and long horizons and is less accurate than
TimesFM-3 at short horizons (Table~\ref{tab:gifteval-extended}). By domain,
among the systems compared, it is the most accurate on Energy and Transport,
which together account for nearly half of the configurations
(Table~\ref{tab:app-domain}).

\section{Ablations and Analysis}\label{sec:ablations}\label{sec:analysis}

We examine the four design choices of
Sections~\ref{sec:method-tla}--\ref{sec:method-multiscale} together with the
pretraining recipe of Section~\ref{sec:setup-data}. Each comparison is a
controlled study that varies only the factor under test, and significance
statements use 95\% paired bootstrap intervals resampled over
configurations~\citep{efron1979bootstrap}.

\subsection{Dual-Axis Attention}\label{sec:ablation-tla}

Because the grid is already serialized into one token sequence, unrestricted
attention over all pairs is available as a baseline. Table~\ref{tab:abl-tla}
compares three nine-layer encoders matched in width, patch width, and every other
architectural and training setting: full attention, dual-axis attention whose
two pathways are averaged with equal weight, and dual-axis attention with the
learned gate.

\begin{table}[t]
\centering
\small
\caption{Attention-operator comparison. All metrics are relative; lower is
better.}
\label{tab:abl-tla}
\setlength{\tabcolsep}{6.0pt}
\begin{tabular}{l r r r}
\toprule
Attention operator & MASE & CRPS & Multivariate MASE \\
\midrule
Full attention over the grid & 0.759 & 0.567 & 0.774 \\
Dual-axis attention, fixed gate & 0.725 & 0.550 & 0.748 \\
\textbf{Dual-axis attention, learned gate} & \textbf{0.685} & \textbf{0.519} & \textbf{0.703} \\
\bottomrule
\end{tabular}
\end{table}

Dual-axis attention is the strongest operator on all three measures, reducing
error against full attention by $9.7\%$ in MASE, $8.4\%$ in CRPS, and $9.1\%$ on
multivariate configurations. Restricting interactions to the two grid axes is
therefore a useful inductive bias rather than a loss of expressiveness: full
attention can represent the same relations but must learn them from data, which
it does less effectively at a matched training budget. The fixed-gate variant sits between the two, so the gain
splits: restricting attention to the grid axes is worth $4.5\%$ and learning the
balance a further $5.5\%$. The larger share comes from the learned gate, nine
scalars that let the two pathways contribute unequally at each depth. Dual-axis attention
also cuts peak activation memory by more than half, allowing larger per-device
batches (\appref{architecture}{A1}).

\subsection{Axis-Aware Two-Dimensional RoPE}
\label{sec:ablation-position}

We compare four arms under one dual-axis encoder on the same data: no
positional encoding, the flattened index $p=tN+n$, temporal-only RoPE with a
learned variable bias, and axis-aware two-dimensional RoPE. Removing positional
information costs $85\%$ relative MASE, far beyond any difference among the
encoded arms, so recovering the grid geometry is the dominant effect.
Axis-aware RoPE is strongest, but its margin over temporal-only RoPE is not
significant on aggregate GIFT-Eval, which is dominated by univariate and
narrowly multivariate series, while on targeted suites it leads by $1.4\%$ on
covariate-driven series and by $9.0\%$ on schema widths held out from
training, both significant. Explicit variable geometry therefore earns its
place where the variable axis carries information
(\appref{position-encoding}{D1}).

\subsection{Horizon Scalability}
\label{sec:ablation-horizon}

We compare four patch-width-two encoders at matched capacity and update budget,
differing only in the horizon objective: fixed blocks of 1024 and of 512
timesteps, lead-time reweighting, which supervises the wide block under a
weight decaying with lead time, and horizon sampling from the curriculum of
\appref{prediction-curriculum}{C1}. A fixed block is correct for one span only,
so neither fixed arm wins: the wide one never supervises short requests on
their own, and halving it improves short horizons by $4.5\%$ but pushes medium
and long spans into iterative rollout, losing more than it gains. Reweighting
the wide block toward early lead times recovers too little to cover its
long-range cost and ends level with both. Horizon sampling is the best of the
four and the only arm below both fixed baselines; reweighting the loss toward
short horizons is insufficient, and the model must be supervised on them
directly (\appref{horizon-objectives}{D2}). The released
model beats every arm on medium and long horizons and trails on short, the gap
Section~\ref{sec:ablation-multiscale} narrows.

\subsection{Multiresolution Forecasting}
\label{sec:ablation-multiscale}

Different patch widths resolve different kinds of structure.
Table~\ref{tab:abl-synthetic} scores all three on synthetic families grouped by
measured signal statistics (\appref{resolution-study}{D3}), and none wins more
than half. Width 2 wins where the informative structure is finer than a wide
token, in series that oscillate within it and in
wide schemas that leave each variable a
short span. Width 4 wins on the event-driven families, wide enough to describe
an event yet narrow enough to keep two apart. Width 8 wins on
regime-persistent levels, reaching back far enough to see the switches.

\begin{table}[t]
\centering
\small
\caption{Resolution study on synthetic families, relative MASE; best single
width in bold. Families are defined by measured signal statistics. Average
pools the single-width encoders with equal weights; Multiscale pools the same
encoders under the learned mixing head.}
\label{tab:abl-synthetic}
\setlength{\tabcolsep}{5.0pt}
\begin{tabular}{l r r r r r}
\toprule
Signal family & Patch 2 & Patch 4 & Patch 8 & Average & Multiscale \\
\midrule
High-frequency oscillations & \textbf{0.888} & 0.895 & 0.976 & 0.875 & 0.870 \\
Wide multivariate schemas & \textbf{0.405} & 0.435 & 0.427 & 0.387 & 0.379 \\
Sparse impulses & 0.873 & \textbf{0.815} & 0.970 & 0.851 & 0.809 \\
Volatility bursts & 0.644 & \textbf{0.557} & 0.776 & 0.586 & 0.563 \\
Low signal-to-noise & 0.702 & \textbf{0.473} & 0.711 & 0.602 & 0.473 \\
Regime-persistent levels & 0.531 & 0.530 & \textbf{0.486} & 0.497 & 0.484 \\
\bottomrule
\end{tabular}
\end{table}

Effective combination must therefore be instance-adaptive, which is what the
mixing head of Section~\ref{sec:method-multiscale} supplies. Each family pools
many series, so a weighting that varies within a family can fall below every
fixed width. The last two columns separate adaptation from the mere presence of
several resolutions. The learned head beats an equal-weight blend of the same
encoders on every family, and matches or beats the best single width on all but
one. The same holds on GIFT-Eval (\appref{mixing-controls}{D4}): a single
patch-4 encoder grown to 247M parameters is $22.3\%$ worse than \morpheustwo{} at 81.5M,
whereas pooling improves on the strongest of its five encoders by $6.8\%$ in
MASE and $9.0\%$ in CRPS. Patch widths coarser than 8 add no information the
retained set lacks.

\subsection{Synthetic Data and Training Curriculum}\label{sec:ablation-data}

Observed series are hard to learn from directly: periodicity, trend, and
cross-variable coupling arrive entangled and buried under measurement noise and
missingness, so a transformer trained on real data alone must acquire every
pattern at once. A curriculum avoids that but needs simpler material to start
from, which is what MySTiC supplies: its underlying factors are specified, so
each pattern can be presented in isolation and free of
that irregularity, and training can run from purely synthetic to predominantly
real. Three parameter-matched arms isolate the two contributions
(\appref{data-curriculum}{D5}): adding MySTiC as an undifferentiated mixture
improves relative MASE by $6.7\%$, and ordering the same corpus into four
stages yields a further $9.7\%$, $15.8\%$ in total.

\section{Limitations and Future Work}\label{sec:limits}

Accuracy is not uniform across the benchmark. Among the systems compared by
domain, \morpheusmultiscale{} leads Energy, the largest domain, and Transport,
and trails on Nature and Healthcare; the probabilistic gap concentrates in
Healthcare, Nature, and Web/CloudOps. Pooling brings nominal $80\%$ intervals to
$75.7\%$ coverage, from $49.3\%$ for the standalone model
(\appref{calibration}{E4}). Three properties of the method
bound these results: requests beyond the native block are reached by iterative
rollout and accumulate error, the fixed token budget of the interleaved
canvas trades context depth for the number of variables, and running five
encoders makes \morpheusmultiscale{} costlier at inference than the standalone
model.

Each points to a direction. The patch widths are fixed by hand, where the
granularity a series needs could instead be learned during training; removing
the rollout penalty means supervising the model on context it generated itself;
and modeling dependence across targets and lead times would replace the marginal
quantile grid with a joint predictive distribution. The mixing head is the most
portable of the three: it reads only the context and horizon, never the
components' forecasts, and trains from their errors, so a component need only
emit aligned quantiles and can otherwise stay fixed or opaque. Alternative
backbones~\citep{tirex,stride} and published systems~\citep{chronos2,timesfm3}
all meet that interface, and
\appref{gifteval-extended}{E1} shows where the gain would be, since our weakness
is concentrated at short horizons.

\section{Conclusion}\label{sec:conclusion}

\morpheustwo{} treats the three ways forecasting tasks differ, in schema,
horizon, and temporal resolution, as modeling problems rather than
configuration choices. Dual-axis attention relates variables to one another and
each variable to its own history inside every layer, with a learned gate
setting their balance at each depth; prediction-length randomization trains one parameter set
across horizons spanning two orders of magnitude at no inference cost; and
resolving a series at several patch widths under a context-conditioned
weighting improves both point and probabilistic accuracy over the standalone
model. The two
releases give complementary parameter budgets, an 81M-parameter model
competitive with systems several times its size and a larger one that attains
the lowest relative MASE and CRPS of any published system on GIFT-Eval, with efficiency
claimed as parameter count at matched accuracy rather than latency or memory.
All three mechanisms are portable, asking only that a model keep both grid
coordinates, sample its training horizon, or emit aligned quantiles, and we
offer them as design principles for future time-series foundation models.


\clearpage
\nocite{gneiting2007,krell2021packing,raffel2020t5,rasmussen2006gp}
\bibliographystyle{morpheus2}
\bibliography{references}

\clearpage

\newcommand{\suppanchor}[1]{\hypertarget{supp-#1}{}}
\makeatletter
\renewcommand\paragraph{\@startsection{paragraph}{4}{\z@}%
  {0.6ex \@plus .2ex \@minus .1ex}{-0.6em}%
  {\normalfont\normalsize\bfseries}}
\makeatother

\appendix
\renewcommand{\thesection}{\Alph{section}}
\renewcommand{\thesubsection}{\thesection\arabic{subsection}}

\begin{center}
{\Large\bfseries Appendix}
\end{center}
\vspace{0.5em}

\suppanchor{training}
\section{Implementation Details}\label{app:training}
This section gives the implementation details omitted from
\mainref{sec:method}{4}, in the same order: the token layout and how the two attention neighborhoods are
executed (Section~\ref{app:architecture}), the positional encoding
(Section~\ref{app:position-impl}), the configuration of every reported model
(Section~\ref{app:model-manifests}), and the packing and optimization settings
(Section~\ref{app:optimization}).

\suppanchor{architecture}
\subsection{Encoder Architecture}\label{app:architecture}
\mainref{sec:method-tla}{4.3} defines the encoder and its two attention
neighborhoods. Rows are ordered by variable type, past-only covariates first,
then known-future covariates, then targets, before time-major interleaving, and
padding is excluded from both neighborhoods. Block-sparse
FlexAttention~\citep{flexattention} executes the two neighborhoods without
materializing the discarded pairs; unrestricted attention instead more than
doubles peak activation memory, so the operator comparison matches effective
batch size with gradient accumulation. The fusion gate is one scalar per layer,
initialized so that both pathways start at weight $0.5$ and learned from the
forecasting loss alone. Requests longer than the native block are served by
iterative rollout: known-future covariates keep their supplied values, past-only
covariates are marked unobserved, and the predicted median is fed back, so
uncertainty does not propagate across passes.

\suppanchor{position-impl}
\subsection{Positional Encoding}
\label{app:position-impl}
\mainref{sec:method-position}{4.4} defines the rotary scheme. The two
coordinates cannot alias because each pathway holds one fixed: the intra-time
pathway compares tokens with $\Delta t=0$ and the inter-time pathway tokens with
$\Delta n=0$. The variable-axis scale $\alpha=0.04$ and the rotary base
$\theta=5000$ are fixed by grid search on a held-out multivariate synthetic
suite; $\alpha$ must be small enough that one row of displacement costs far less
phase than one timestep, and $\theta$ balances local resolution against reach
over the context lengths we train on.

\suppanchor{model-manifests}
\subsection{Model Configurations and Hardware}
\label{app:model-manifests}
\mainref{sec:setup-training}{5.2} states that every reported model shares one
encoder specification. Table~\ref{tab:app-encoder-spec} gives it.

\begin{table}[h]
\centering
\caption{Shared encoder specification. Patch width, the token budget $P$, and
training hyperparameters vary across the reported models.}
\label{tab:app-encoder-spec}
\small
\setlength{\tabcolsep}{6.0pt}
\begin{tabular}{l l}
\toprule
Field & Value \\
\midrule
Layers                   & 9 dual-axis blocks \\
Hidden width             & 640 \\
Attention heads          & 20 \\
Head dimension           & 32 \\
Feed-forward width       & 4096 \\
Rotary base $\theta$     & 5000 \\
Variable-axis scale $\alpha$ & 0.04 \\
Normalization            & context-only per row, arcsinh-stabilized \\
Quantile levels          & 21 (11 for two \morpheusmultiscale{} encoders) \\
Max patches              & $P$ \\
Max prediction patches   & $P/32$ \\
\bottomrule
\end{tabular}
\end{table}

\noindent
Standalone \morpheustwo{} sets patch width 4 and $P=8{,}192$, giving 81.5M
parameters. \morpheusmultiscale{} combines five independently trained instances
of this specification, two at patch width 2, two at width 4, and one at width 8,
which differ in training hyperparameters; token budgets scale with width so that
every encoder covers the same context span and native horizon ($P=16{,}384$,
$8{,}192$, and $4{,}096$ at widths 2, 4, and 8, giving 31,744 timesteps of context
and a 1,024-step native horizon for a univariate series). With the
1.4M-parameter mixing head the system has 413M parameters. All reported models
are trained on a single eight-GPU A100-40GB node with standard distributed data
parallelism.

\suppanchor{optimization}
\subsection{Sequence Packing and Optimization}
\label{app:optimization}
\mainref{sec:setup-training}{5.2} summarizes optimization. Forecasting examples
vary widely in length and schema width, so several are packed into one
interleaved canvas~\citep{raffel2020t5,krell2021packing}; each carries its own
sample identifier, which the attention masks require to match, so packing
changes throughput without changing any model output. The packer stages 32
examples, keeps at most 16 canvases open, and releases a canvas at $85\%$
occupancy. Training uses
AdamW~\citep{adamw} in bfloat16 under a one-cycle
schedule~\citep{smith2018onecycle}; default settings are four packed sequences
per device, peak learning rate $10^{-3}$ with a $10\%$ warm-up, cosine decay to
$4\times10^{-4}$, no weight decay, and gradient clipping at 10. Because a packed
sequence holds a
variable number of examples, training budgets are compared by realized
patch-token exposure per trainable parameter~\citep{hoffmann2022training}.

\suppanchor{data}
\section{Dataset Details}\label{app:data}

\suppanchor{training-corpora}
\subsection{Pretraining Corpora}
\label{app:training-corpora}
\mainref{sec:setup-data}{5.1} introduces the real portion of the mixture:
GIFT-Eval Pretrain, the public pretraining corpus released with the
benchmark~\citep{gifteval}, which excludes its test windows.
Table~\ref{tab:app-pretrain-domains} reports its composition.

\begin{table}[h]
\centering
\caption{Composition of GIFT-Eval Pretrain. Series counts target rows, not
items; size is on-disk Arrow. Domain
assignment follows the GIFT-Eval taxonomy, and every partition is assigned to
exactly one domain.}
\label{tab:app-pretrain-domains}
\small
\setlength{\tabcolsep}{5.0pt}
\begin{tabular}{l r r r r l}
\toprule
Domain & Partitions & Series & Obs. (B) & Size (GB) & Frequencies \\
\midrule
Nature        &  70 & 25,397,091 & 204.5 & 818.2 & H, 6H, D \\
Energy        &  29 &  1,804,428 &  16.9 &  68.9 & 4S, 1min, 10min, 30min, H \\
Transport     &  24 &    165,266 &   4.9 &  19.6 & 5min, 15min, 30min, H, D, W \\
Web/CloudOps  &   6 &    900,863 &   2.6 &  30.5 & 5min, D, W \\
Sales         &   4 &    148,654 &   0.2 &   0.8 & D, M \\
Healthcare    &   3 &      1,929 & $<0.1$ & $<0.1$ & W \\
Econ/Fin      &  16 &      5,555 & $<0.1$ & $<0.1$ & D, W, M, Q, Y \\
\midrule
\textbf{Total} & \textbf{152} & \textbf{28,423,786} & \textbf{229.1} &
  \textbf{938.0} & 13 distinct \\
\bottomrule
\end{tabular}
\end{table}

The corpus is severely imbalanced: climate reanalysis accounts for $87\%$ of
bytes and $89\%$ of series, whereas Healthcare and Econ/Fin together contribute
under $0.1\%$. We therefore assign each domain--frequency combination its own
sampling weight, so that no single source dominates a batch and the smallest
still appear regularly.

\suppanchor{mystic}
\subsection{Synthetic Data Generation}
\label{app:mystic}

\mainref{sec:setup-data}{5.1} introduces MySTiC. Public corpora concentrate in a
few sources and frequencies (Section~\ref{app:training-corpora}), so wide panels,
labelled covariate relations, and controlled nonstationarity are thinly
represented. MySTiC fills these gaps with generated series whose factors are
specified, not inferred, so each kind of temporal structure can be presented on
its own, free of the measurement noise and missingness of observed data. This
gives the early curriculum stages (Section~\ref{app:data-schedule}) material a
model can learn from before it meets real series, which makes training from
random initialization tractable. Every family is forecastable: each carries a
repeating waveform, a persistent level, or a recoverable covariate relation, and
noise is added on top of that structure, not substituted for it. Records use the
model's variable--time layout, and deterministic seed keys make each record
reproducible. Table~\ref{tab:app-mystic-composition} gives the composition of
the three subsets.

\begin{table}[h]
\centering
\caption{MySTiC composition. A record is one multivariate series and rows are
its variables.}
\label{tab:app-mystic-composition}
\small
\setlength{\tabcolsep}{5.0pt}
\begin{tabular}{l r r l r r}
\toprule
Subset & Collections & Records & Rows/record & Obs. (B) & Size (GB) \\
\midrule
Kernel composition   & 42 &   357,000 & 1--21 &  5.7 &  24.1 \\
Covariate response   & 14 &   365,004 & 1--5 targets &  2.3 &  34.9 \\
Real-world emulation & 82 &   585,890 & 1--21 & 14.3 &  61.6 \\
\midrule
\textbf{Total} & \textbf{138} & \textbf{1,307,894} & & \textbf{22.3} & \textbf{120.6} \\
\bottomrule
\end{tabular}
\end{table}

\begin{figure}[t]
\centering
\includegraphics[width=\textwidth]{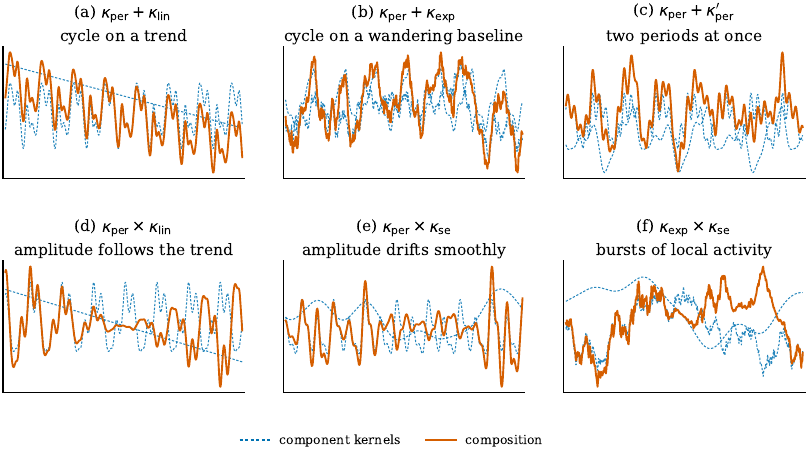}
\caption{Kernel composition in the MySTiC kernel subset. Each row is drawn from
a Gaussian process whose covariance is assembled from base kernels under
sums and products: a sum superposes
independent structure, and a product modulates one component by another. Panels
combine periodic, linear, exponential, and squared-exponential kernels
($\kappa_{\mathrm{per}}$, $\kappa_{\mathrm{lin}}$, $\kappa_{\mathrm{exp}}$,
$\kappa_{\mathrm{se}}$), with the component draws in blue and the composition
they produce in red. Sums give a cycle on a trend (a), a cycle on a wandering
baseline (b), and two periods at once (c); products give periodic amplitude that
follows a trend (d) or drifts smoothly (e), and bursts of local activity (f).
Series are standardized for display.}
\label{fig:mystic-kernel}
\end{figure}

\paragraph{Kernel composition.} This subset varies periodicity, trend, memory,
and cross-row dependence independently, properties that observed panels deliver
bundled with their source. Extending the generator of
Morpheus~\citep{morpheus1}, each row is drawn from a Gaussian-process
prior~\citep{rasmussen2006gp} whose covariance combines periodic, trend,
decaying, and noise kernels under randomly chosen sums and products, evaluated
on a reference-scaled grid so that a configured cycle keeps its period across
series lengths (Figure~\ref{fig:mystic-kernel}). Long series instead specify a
power spectrum directly. Cross-row dependence enters through latent spectral
factors, which give each row a shared and an idiosyncratic component, and
through lead--lag couplings, in which one row follows a driver after a sampled
delay. Level shifts on all or some rows, a slow variance envelope, and per-row
rescaling add nonstationarity and scale variation, so predictive intervals must
adapt over time.

\begin{figure}[t]
\centering
\includegraphics[width=\textwidth]{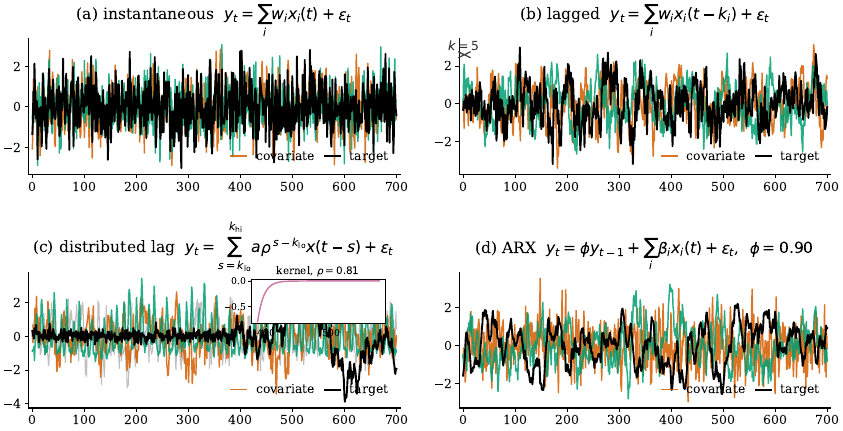}
\caption{Covariate-response subset. Black is the target, colored lines are the
covariates that drive it, and grey lines are distractors that drive nothing; all
rows are standardized for display. (a) instantaneous linear response; (b)
lagged response, with the realized lag $k$ annotated; (c) distributed lag, with
the geometric kernel inset; (d) autoregressive response with covariates, with
its persistence parameter $\phi$ annotated.}
\label{fig:mystic-covariate}
\end{figure}

\paragraph{Covariate response.} Public corpora supply covariate columns without
identifying which matter, at what lag, or how strongly. This subset fixes the
relation by construction, so covariate-conditioned supervision is available and
forecasts can be compared against an oracle with the same inputs. Covariate rows
are standardized draws from periodic, piecewise-constant, or autoregressive
processes. Each target is assigned a driving subset of rows and one of five
response recipes (Figure~\ref{fig:mystic-covariate}): instantaneous linear;
instantaneous nonlinear, with saturating, rectifying, or periodic maps and
multiplicative interactions between drivers; lagged linear; distributed lag
under a geometrically decaying kernel; and autoregressive with exogenous inputs.
The recipes differ in what a forecaster must do: read across rows at one time
index, align rows in time, integrate a driver over a window, or track the
target's own persistence. Coefficients have a nonzero magnitude floor and noise
is scaled to the realized signal, so signal-to-noise is comparable across
recipes. Known-future rows remain visible over the horizon while past-only rows
stop at the forecast origin, so a past-only driver helps only when its lag
places the needed values in context. Distractor rows drawn from the same
processes drive nothing, so marginal appearance cannot identify a relevant
input.

\begin{figure}[t]
\centering
\includegraphics[width=\textwidth]{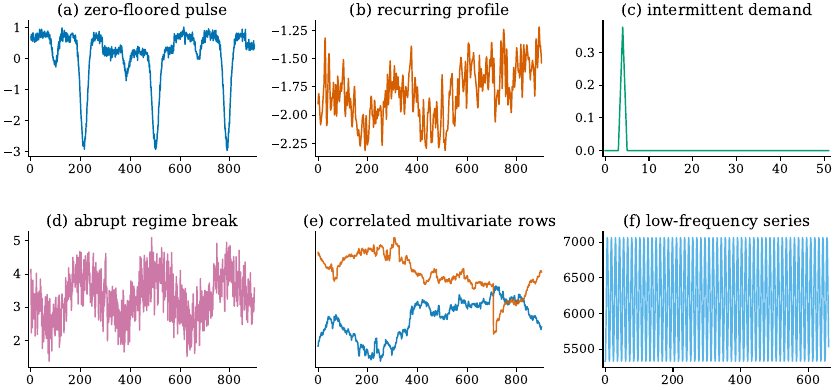}
\caption{Real-world emulation subset. (a) zero-floored pulse from the
trapezoid primitive, with exact zeros between cycles;
(b) a recurring profile under amplitude modulation; (c) a short intermittent
record; (d) an abrupt
regime break; (e) a correlated multivariate record, rows offset vertically for
legibility; (f) a low-frequency series. Panels are individual records.}
\label{fig:mystic-emulation}
\end{figure}

\paragraph{Real-world emulation.} This subset supplies a repeating conditional
mean, the structure seasonal baselines rely on. Periods are set in wall-clock
time, so one profile renders at any sampling frequency and at many ratios of
cycle length to patch width. Within a period, trapezoid and bump primitives give
pulses with exact zeros, plateaus, and smooth humps; across periods, a
persistent amplitude, a weekly modulation, and a piecewise-constant activity
state vary the profile. Trend, noise, spikes, and regime breaks placed inside
forecast windows complete the record (Figure~\ref{fig:mystic-emulation}).

\suppanchor{data-schedule}
\subsection{Data Mixture and Curriculum Schedule}
\label{app:data-schedule}
The mixture is presented in four stages of increasing realism. The first, $15\%$
of the update budget, is dominated by the kernel-composition subset; the second,
$20\%$, mixes all three synthetic subsets with a little GIFT-Eval Pretrain; the
third is roughly half real; and the fourth is about $90\%$ real. Within the real
portion every stage keeps the domain--frequency weights of
Section~\ref{app:training-corpora}.

\suppanchor{protocols}
\section{Training and Evaluation Protocols}\label{app:protocols}

\suppanchor{prediction-curriculum}
\subsection{Prediction-Length Curriculum}
\label{app:prediction-curriculum}
Every reported encoder is trained with prediction-length randomization. At patch
width 2 the curriculum samples
\begin{equation}
\{m_k\}=\{4,6,15,24,240,360,512\},\qquad
\{\pi_k\}=\{.04,.03,.06,.12,.20,.20,.35\},
\end{equation}
and widths 4 and 8 use the same probabilities over budgets scaled by $2/s$, so
sampled horizons in timesteps match across widths. One $M$ is shared by every
example in a batch, which preserves compatible packed layouts.

\suppanchor{scale-training}
\subsection{Mixing-Head Training}
\label{app:scale-training}
\mainref{sec:method-multiscale}{4.6} defines the pool. The mixing head adds
1.4M parameters, about $0.3\%$ of the \morpheusmultiscale{} inference budget,
and never reads the encoders' forecasts, so its weights depend on the request
alone.

\begin{figure}[t]
\centering
\includegraphics[width=\textwidth]{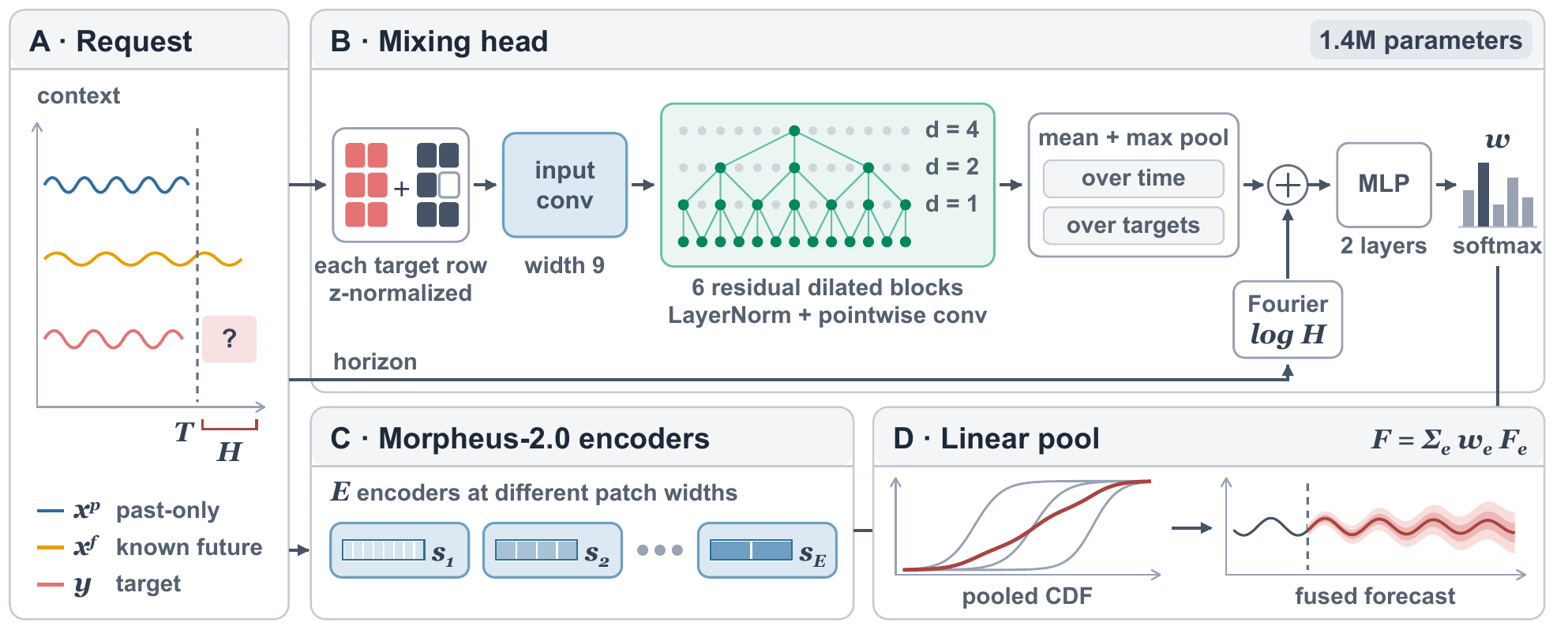}
\caption{Mixing head of \morpheusmultiscale{}. \textbf{(A)} A request is the
context up to the forecast origin $T$ and the horizon $H$. \textbf{(B)} Each
target row is z-normalized, stacked with its observation mask, and encoded by an
input convolution and six residual dilated-convolution blocks; masked mean and
maximum pooling over time and then over targets give a fixed-width summary,
which is concatenated with a Fourier embedding of $\log H$ and mapped by a
two-layer classifier to one weight per encoder. \textbf{(C)} $E$ \morpheustwo{}
encoders, trained at different patch widths, each forecast the
same request. \textbf{(D)} The forecast is the weighted linear pool of their
predictive distributions. The head never reads the encoders' forecasts, so its
weights depend on the request alone.}
\label{fig:mixing-head}
\end{figure}

\paragraph{Architecture.} Each target row of the context, up to its most recent
32,768 steps, is z-normalized with statistics from its observed values and
stacked with its observation mask. A shared stack processes every row
independently: an input convolution of width 9 followed by six residual blocks
of 192 channels, each a dilated convolution with dilation $1,2,4,\ldots,32$,
layer normalization, and a pointwise convolution. Masked mean and maximum
pooling over time give one vector per target, and a second masked mean and
maximum over targets give a fixed-width summary invariant to the number and
order of targets. A Fourier embedding of the log horizon is concatenated, and a
two-layer classifier of width 384 emits one logit per encoder
(Figure~\ref{fig:mixing-head}).

\paragraph{Objective.} The head is trained on windows from the GIFT-Eval
training split, on which every encoder's forecast is precomputed. For window
$i$, let $s_{i,e}$ be the quantile loss of encoder $e$ summed over lead times and
the nine benchmark levels, and $\bar s_i$ its mean over encoders. The head's
softmax output $p_i$ is fitted to a soft target by weighted cross-entropy,
\begin{equation}
q_{i,e}=\frac{\exp\!\bigl(-s_{i,e}/(T\bar s_i)\bigr)}
             {\sum_{e'}\exp\!\bigl(-s_{i,e'}/(T\bar s_i)\bigr)},
\qquad
\mathcal L=-\sum_i r_i\sum_e q_{i,e}\log p_{i,e},
\end{equation}
with a small temperature $T$. Dividing by $\bar s_i$ makes the target
scale-free, so a few large-magnitude windows cannot dominate. Near-ties give a
diffuse target that teaches the head to hedge, and large gaps a near one-hot
target that teaches it to commit. The weight $r_i$ is proportional to the
relative regret $(\bar s_i-\min_e s_{i,e})/\bar s_i$, so windows on which the
choice of encoder barely matters carry little weight, and is balanced across
datasets. Optimization uses AdamW with warm-up and cosine decay.

\paragraph{Pooling.} At inference the head's output is mixed in equal parts
with uniform weights, which keeps every encoder in the pool and guards against
overconfident routing. The pool is formed on the nine benchmark levels: each
encoder's CDF is interpolated linearly between its quantiles at
$0.1,\ldots,0.9$, and beyond its outer quantiles is known only to lie in
$[0,0.1]$ or $[0.9,1]$. The reported quantile is the midpoint of the pools
obtained at the two extremes, so no quantile outside an encoder's grid is
extrapolated.

\suppanchor{quantile-evaluation}
\subsection{Metric Definitions and Aggregation}
\label{app:quantile-evaluation}
This section defines precisely the metrics named in
\mainref{sec:setup-metrics}{5.3}.
Most encoders emit the 21-level grid
\begin{equation}
\mathcal Q=\{.01,.05,.10,.15,\ldots,.90,.95,.99\};
\end{equation}
two \morpheusmultiscale{} encoders emit the 11-level grid
$\{.01,.1,.2,\ldots,.9,.99\}$. Both contain the nine benchmark levels.
For target $j$ and lead time $h$, the model estimates the conditional quantile
function
\begin{equation}
Q_{j,h}(\tau\mid\mathcal C_T)=\inf\{q:\Pr(y_{j,T+h}\le q\mid\mathcal C_T)\ge\tau\},
\end{equation}
the one-dimensional marginal of the conditional law of $\mathbf y_{T+1:T+H}$
given $\mathcal C_T$.
The implementation multiplies the conventional pinball
loss~\citep{koenker1978} by two, a constant
that does not change its minimizer. It does not impose a non-crossing
constraint, and its outputs are marginal quantiles for each target and lead
time rather than a joint multivariate distribution. Evaluation linearly
interpolates native levels onto the benchmark grid when necessary. Requests
longer than the native block are completed by iterative rollout, feeding back
the predicted medians and retaining the quantiles of each pass.

Benchmark scoring uses the 9-level grid $\mathcal Q_G=\{0.1,0.2,\ldots,0.9\}$
\citep{gifteval}.
For configuration $c$, let $\Omega_c$ contain all valid item--target--horizon
values, and let $i$ index an item--target pair, the benchmark's counterpart to
the target index $j$ of the main paper's problem formulation, with history
length $T_i$, seasonal period $m_i$, and historical seasonal scale
$\mathrm{sn}_i$. The two headline metrics are
\begin{align}
\mathrm{MASE}_c
 &=\frac{1}{|\Omega_c|}\sum_{(i,h)\in\Omega_c}
   \frac{|y_{i,h}-\widehat Q_{i,h}(0.5)|}{\mathrm{sn}_i},
&\mathrm{sn}_i&=\frac{1}{T_i-m_i}\sum_{t=m_i+1}^{T_i}|y_{i,t}-y_{i,t-m_i}|,\\
\mathrm{MWQL}_c
 &=\frac{1}{9}\sum_{\tau\in\mathcal Q_G}
   \frac{2\sum_{(i,h)\in\Omega_c}\rho_\tau(y_{i,h}-\widehat Q_{i,h}(\tau))}
         {\sum_{(i,h)\in\Omega_c}|y_{i,h}|},
&\rho_\tau(u)&=\max\{\tau u,(\tau-1)u\}.
\label{eq:gifteval-metrics}
\end{align}
The continuous ranked probability score (CRPS) rewards a predictive distribution
for concentrating near the observed value. The GIFT-Eval quantity called CRPS is
operationally its normalized aggregate of
mean weighted quantile loss, rather than a numerical integral of a continuous
predictive CDF. Headline metrics use all evaluation items and geometric means
over the 97 configurations. Writing $L_c$ for a metric on configuration $c$, the
reported relative value is
\begin{equation}
\mathrm{Rel}(L)=
\frac{\exp\!\left(\frac{1}{97}\sum_c\log L_c\right)}
      {\exp\!\left(\frac{1}{97}\sum_c\log L_c^{\mathrm{SeasonalNaive}}\right)},
\label{eq:relative-gifteval}
\end{equation}
so Seasonal Naive scores $1.0$ by construction. The reference predictor repeats
the observation from one native-frequency seasonal period earlier. Its
unnormalized geometric means over the 97 configurations are $1.398$ for MASE and
$0.252$ for MWQL, which recover absolute values from any relative figure
reported in the paper.
Within each configuration, GluonTS~\citep{gluonts} aggregates MASE over all
valid item, target,
and lead-time values; MWQL first sums quantile losses and absolute labels over
the same values and then forms their ratio. Seasonal periods are inferred from
the native frequency. If the nominal period exceeds the available history,
GluonTS uses a lag-one scale for MASE~\citep{hyndman2006mase}.

\suppanchor{setup-categories}
\subsection{Evaluation Setup Categories}
\label{app:setup-categories}
This section defines the Setup column used in the result tables of
\mainref{sec:results}{6}.
Submissions are distinguished by which GIFT-Eval data a model was exposed to
before scoring, and we use three labels throughout. Zero-shot models are trained
without any GIFT-Eval data, including its training split. Pretrained models may
include the GIFT-Eval training split in their pretraining mixture and are then
scored with one set of weights across all configurations. Fine-tuned models
additionally adapt a pretrained checkpoint to the benchmark training split. All
three categories exclude the GIFT-Eval test windows; the distinction concerns
the training split only.

\morpheustwo{} and \morpheusmultiscale{} are trained on GIFT-Eval Pretrain,
which includes the GIFT-Eval training split, and are scored with one set of
weights across all 97 configurations without per-configuration adaptation. They
are therefore reported as pretrained on the benchmark.
Table~\ref{tab:gifteval-extended} records the category of every system it lists,
taken from that system's own public reporting.

\suppanchor{ablations}
\section{Additional Ablations}\label{app:ablations}

\suppanchor{position-encoding}
\subsection{Positional Encoding Ablations}
\label{app:position-encoding}
This section gives the arms and measurements summarized in
\mainref{sec:ablation-position}{7.2}.
Table~\ref{tab:position-designs} places the arms structurally before
Table~\ref{tab:abl-position} scores them. The two named alternatives are
temporal RoPE with learned variable identifiers~\citep{roformer} and
alternating axial attention~\citep{axialattention,crossformer}.

\begin{table}[h]
\centering
\caption{Structural comparison of positional and attention designs; entries
summarize inductive biases.}
\label{tab:position-designs}
\small
\setlength{\tabcolsep}{4.2pt}
\begin{tabular}{l c c c}
\toprule
Design & Time geometry & Variable geometry & Both axes per layer \\
\midrule
Flattened one-dimensional positions & Entangled with $N$ & Entangled with $N$ & Model dependent \\
Temporal RoPE + variable identifiers & Relative & Absolute & Model dependent \\
Alternating axial attention & Relative & Relative & No \\
\textbf{Ours} & \textbf{Relative} & \textbf{Relative} & \textbf{Yes} \\
\bottomrule
\end{tabular}
\end{table}

All four arms share data and recipe, and differ in parameters only by a variable
embedding worth $0.04\%$ of the total (Table~\ref{tab:abl-position}). The
covariate-informed suite holds 7,000 synthetic series whose target is a known
function of up to 8 rows; the held-out-width suite holds 300 series at each
width from 2 to 21 rows, with widths 12, 16, and 21 excluded from the training
data of these arms. Against
temporal RoPE with a variable bias, axis-aware RoPE leads by $1.4\%$ (95\% CI
$[1.0,1.8]$) and $9.0\%$ ($[7.5,10.5]$) on the two suites, while its $1.4\%$
aggregate margin on GIFT-Eval is not significant.

\begin{table}[h]
\centering
\caption{Positional-encoding error on aggregate, covariate-informed, and
held-out-width suites. Lower is better.}
\label{tab:abl-position}
\small
\setlength{\tabcolsep}{3.8pt}
\begin{tabular}{l r r r r}
\toprule
& \multicolumn{2}{c}{All configurations} & Covariate-informed & Held-out width \\
\cmidrule(lr){2-3}\cmidrule(lr){4-4}\cmidrule(lr){5-5}
Position scheme & Rel. MASE & Rel. CRPS & Rel. MASE & Rel. MASE \\
\midrule
\textbf{Axis-aware two-dimensional RoPE} & \textbf{0.685} & \textbf{0.519} & \textbf{0.703} & \textbf{0.708} \\
Temporal RoPE + variable bias & 0.695 & 0.531 & 0.713 & 0.778 \\
Flattened one-dimensional RoPE & 0.700 & 0.540 & 0.736 & 0.727 \\
No positional encoding & 1.269 & 0.927 & 1.111 & 2.527 \\
\bottomrule
\end{tabular}
\end{table}

\suppanchor{horizon-objectives}
\subsection{Horizon Objectives}
\label{app:horizon-objectives}
Table~\ref{tab:abl-horizon} breaks the four horizon objectives of
\mainref{sec:ablation-horizon}{7.3} down by horizon group; all arms are
patch-width-two encoders at matched capacity and update budget.

\begin{table}[h]
\centering
\caption{Horizon objectives at matched capacity. Entries are relative MASE
within each horizon group; lower is better.}
\label{tab:abl-horizon}
\small
\setlength{\tabcolsep}{5.5pt}
\begin{tabular}{l c r r r r}
\toprule
Horizon objective & Supervised span & Short & Medium & Long & Overall \\
\midrule
Fixed prediction length & 1024 & 0.734 & 0.673 & 0.659 & 0.704 \\
Fixed prediction length & 512 & 0.701 & 0.714 & 0.716 & 0.707 \\
Lead-time loss reweighting & 1024 (weighted) & 0.721 & 0.689 & 0.683 & 0.706 \\
Horizon sampling & 8--1024 & 0.706 & 0.687 & 0.678 & 0.696 \\
\bottomrule
\end{tabular}
\end{table}

\suppanchor{resolution-study}
\subsection{Patch Resolution}
\label{app:resolution-study}
Table~\ref{tab:app-signal-families} defines the families of
\mainref{sec:ablation-multiscale}{7.4}. Each statistic is measured on the target
series before any forecast is scored, every collection belongs to exactly one
family, and all widths are scored on identical windows.

\begin{table}[h]
\centering
\caption{Signal-family definitions for the resolution study. Every statistic is
measured on the target series before any forecast is scored.}
\label{tab:app-signal-families}
\small
\setlength{\tabcolsep}{6.0pt}
\begin{tabular}{l p{0.62\linewidth}}
\toprule
\textbf{Signal family} & \textbf{Defining statistic} \\
\midrule
High-frequency oscillations & Ten-second and five-minute sources; a third of
  spectral energy lies above the band a width-8 patch can represent \\
\midrule
Wide multivariate schemas   & 7 to 21 simultaneous targets \\
\midrule
Sparse impulses             & Differenced kurtosis above 300 at a jump rate
  below $0.04$: isolated spikes on a smooth baseline \\
\midrule
Volatility bursts           & Absolute-difference autocorrelation near $0.3$
  with a jump rate above $0.12$ \\
\midrule
Low signal-to-noise         & Lowest lag-1 autocorrelation and flattest spectra
  in the corpus \\
\midrule
Regime-persistent levels    & Lag-50 autocorrelation between $0.69$ and $0.82$ \\
\bottomrule
\end{tabular}
\end{table}

\suppanchor{mixing-controls}
\subsection{Mixing-Head Controls}
\label{app:mixing-controls}
Two measurements separate the gain of \morpheusmultiscale{} from added capacity.
A single patch-4 encoder made wider and deeper to 247M parameters reaches
$0.838$ relative MASE, $22.3\%$ worse than \morpheustwo{} at 81.5M, so
scaling one encoder does not substitute for combining several.

Pooling, by contrast, improves on every component. The strongest of the five
encoders reaches $0.693$ relative MASE and the strongest $0.500$ relative CRPS,
while the pool reaches $0.646$ and $0.455$, a further $6.8\%$ and $9.0\%$.
Equal weights over the same encoders reach $0.653$ and $0.459$, so the
context-conditioned weights account for the rest.

Table~\ref{tab:app-mixing-ablation} reports an ablation on a smaller
three-encoder system that separates three ways of spending roughly 245M
parameters: one larger encoder, a mixing head over three encoders of the same
patch width, and a mixing head over patch widths 2, 4, and 8. The single larger
encoder is worse than every 80--84M encoder. Combining three patch-4 encoders
improves on the patch-4 encoder by $2.3\%$ in MASE and $5.2\%$ in CRPS, the gain
from combining alone. Giving the three encoders different patch widths improves
a further $2.4\%$ and $3.3\%$, so at matched capacity, mixing different
resolutions outperforms both scaling one encoder and combining encoders of one
resolution.

\begin{table}[h]
\centering
\caption{Isolating the mixing head. Every row uses the same 97 configurations,
evaluation protocol, rollout policy, and benchmark quantile grid, so rows differ
only in what is combined and at what capacity. Lower is better.}
\label{tab:app-mixing-ablation}
\small
\setlength{\tabcolsep}{5.0pt}
\begin{tabular}{l r r r}
\toprule
System & Params & Rel. MASE & Rel. CRPS \\
\midrule
\morpheustwo{}, patch 2                        & 84.3M & 0.696 & 0.507 \\
\morpheustwo{}, patch 4                        & 81.5M & 0.685 & 0.519 \\
\morpheustwo{}, patch 8                        & 80.1M & 0.708 & 0.526 \\
\midrule
Single encoder at multiscale budget, patch 4   & 247M & 0.838 & 0.602 \\
Mixing head over three encoders, all patch 4   & 245M & 0.669 & 0.492 \\
\midrule
\textbf{Mixing head over patch widths 2, 4, 8}
  & 247M & \textbf{0.653} & \textbf{0.476} \\
\bottomrule
\end{tabular}
\end{table}

\suppanchor{data-curriculum}
\subsection{Synthetic Data and Curriculum}
\label{app:data-curriculum}
Table~\ref{tab:abl-data} gives the three parameter-matched arms of
\mainref{sec:ablation-data}{7.5}, which share every training setting except the
pretraining mixture and its order. Rows two and three share generators and
differ only in scheduling, so the larger gain comes from ordering, not new data.

\begin{table}[h]
\centering
\caption{Contribution of the synthetic corpus and the curriculum that orders
it. Entries are relative MASE over the 97 configurations; lower is better, and
the last row is the full recipe.}
\label{tab:abl-data}
\small
\setlength{\tabcolsep}{6.0pt}
\begin{tabular}{l r}
\toprule
Pretraining recipe & Relative MASE \\
\midrule
Real corpus only & 0.814 \\
$+$ MySTiC, single-stage mixture & 0.759 \\
\textbf{$+$ four-stage curriculum} & \textbf{0.685} \\
\bottomrule
\end{tabular}
\end{table}

\suppanchor{results}
\section{Additional Results}\label{app:results}

Every \morpheustwo{} and \morpheusmultiscale{} entry in this section is computed
from the released checkpoints; all other entries are the values their authors
report.

\suppanchor{gifteval-extended}
\subsection{Extended Comparison and Horizon-Wise Accuracy}
\label{app:gifteval-extended}

Table~\ref{tab:gifteval-extended} extends the comparison of
\mainref{sec:results}{6} to a wider
set of model families and decomposes each one by horizon group. Raw per-group
errors are not comparable across groups, so every entry is normalized by the
Seasonal Naive geometric mean computed within its own group. Unnormalized
geometric means are omitted because they differ from the relative values only by
those fixed denominators.

The decomposition separates profiles that the aggregate hides, and the models
divide into two families. Chronos-2~\citep{chronos2} is strongest at short
horizons and degrades
steadily as the horizon grows, ending $13.5\%$ worse at long than at short;
Toto-2.0-2.5B, TimesFM-2.5, and Moirai-2.0-R-Small
\citep{toto2,timesfm,moirai} follow the same shape.
\morpheustwo{} runs the other way, improving from short to long, so it and
Chronos-2 are nearly tied in aggregate while disagreeing at every individual
horizon: Chronos-2 leads by $10\%$ on short, and \morpheustwo{} leads by $20\%$
on long. We attribute this pattern to patch width. A width-4 token resolves less
local detail than Chronos-2 retains, which costs short horizons, but reaches
further into history per token, which pays at long ones. Falcon-X is the only
other entry with that orientation.

\morpheusmultiscale{} improves on \morpheustwo{} in every group and on both
metrics, by the widest margin at short horizons. It records the best relative
MASE and relative CRPS in the table overall and the best of both at medium and
long horizons, where its lead is largest: at long horizons it improves on
TimesFM-3 by $16.7\%$ in MASE and $16.6\%$ in CRPS, and the runner-up at that
horizon on both metrics is our standalone model \morpheustwo{}.

The spread across horizon groups follows from patch width. A token spanning two
or more timesteps reaches further into history but resolves less local detail,
and short requests depend on that detail. Pooling recovers part of the loss,
improving on \morpheustwo{} by $7.4\%$ relative MASE at short horizons, its
largest gain in any group. A deficit against TimesFM-3~\citep{timesfm3}, contemporaneous
work and the strongest prior system on CRPS, remains at short horizons; our
margins at medium and long offset it, leaving a net $3.1\%$. Short requests are
55 of the 97 configurations and carry $57\%$ of the weight in the geometric
mean, so the aggregate margin holds under a weighting that favours the group
where we are weaker.

\begin{table}[t]
\centering
\caption{GIFT-Eval accuracy over 97 configurations, aggregate and by horizon
group. Entries are normalized by Seasonal Naive within the corresponding group
(55 short, 21 medium, 21 long); lower is better. Bold
marks the best entry in each column.}
\label{tab:gifteval-extended}
\footnotesize
\setlength{\tabcolsep}{3.2pt}
\begin{tabular}{l l rr rr rr rr}
\toprule
& & \multicolumn{2}{c}{All} & \multicolumn{2}{c}{Short} & \multicolumn{2}{c}{Medium} & \multicolumn{2}{c}{Long} \\
\cmidrule(lr){3-4}\cmidrule(lr){5-6}\cmidrule(lr){7-8}\cmidrule(lr){9-10}
Model & Setup & MASE & CRPS & MASE & CRPS & MASE & CRPS & MASE & CRPS \\
\midrule
Seasonal Naive & Reference & 1.000 & 1.000 & 1.000 & 1.000 & 1.000 & 1.000 & 1.000 & 1.000 \\
Sundial-Base-128M & Zero-shot & 0.750 & 0.559 & 0.732 & 0.592 & 0.764 & 0.526 & 0.785 & 0.510 \\
Moirai-2.0-R-Small & Pretrained & 0.728 & 0.516 & 0.695 & 0.517 & 0.759 & 0.519 & 0.789 & 0.513 \\
TimesFM-2.5 & Zero-shot & 0.705 & 0.490 & 0.681 & 0.504 & 0.724 & 0.472 & 0.751 & 0.475 \\
Granite-FlowState-r1.1 & Zero-shot & 0.701 & 0.490 & 0.680 & 0.507 & 0.722 & 0.474 & 0.739 & 0.464 \\
Chronos-2 & Pretrained & 0.698 & 0.485 & 0.667 & 0.496 & 0.725 & 0.471 & 0.757 & 0.472 \\
TiRex-2-Zeroshot & Zero-shot & 0.697 & 0.478 & 0.679 & 0.500 & 0.704 & 0.450 & 0.739 & 0.451 \\
Toto-2.0-2.5B & Pretrained & 0.696 & 0.476 & 0.663 & 0.486 & 0.726 & 0.464 & 0.754 & 0.461 \\
Timer-S1 & Pretrained & 0.693 & 0.485 & 0.683 & 0.499 & 0.705 & 0.469 & 0.709 & 0.467 \\
Falcon-X & Pretrained & 0.687 & 0.486 & 0.704 & 0.529 & 0.664 & 0.437 & 0.668 & 0.431 \\
\textbf{\morpheustwo{}} & Pretrained & 0.685 & 0.519 & 0.743 & 0.608 & 0.630 & 0.446 & 0.604 & 0.401 \\
Granite-PatchTST-FM-r2 & Zero-shot & 0.685 & 0.467 & 0.665 & 0.489 & 0.700 & 0.439 & 0.723 & 0.441 \\
Toto-2.0-2.5B-FT & Fine-tuned & 0.679 & 0.463 & 0.653 & 0.476 & 0.706 & 0.450 & 0.724 & 0.444 \\
TiRex-2-Pretrained & Pretrained & 0.678 & 0.467 & 0.662 & 0.489 & 0.692 & 0.443 & 0.706 & 0.436 \\
Toto-2.0-FnF & Pretrained & 0.676 & 0.463 & 0.650 & 0.479 & 0.707 & 0.449 & 0.717 & 0.439 \\
EXAONE-Forecast & Zero-shot & 0.673 & 0.460 & 0.660 & 0.484 & 0.680 & 0.430 & 0.702 & 0.431 \\
TimesFM-3 & Zero-shot & 0.667 & 0.456 & \textbf{0.648} & \textbf{0.475} & 0.685 & 0.432 & 0.700 & 0.430 \\
Falcon-2.0 & Pretrained & 0.666 & 0.486 & 0.652 & 0.492 & 0.667 & 0.474 & 0.703 & 0.483 \\
\textbf{\morpheusmultiscale{}} & Pretrained & \textbf{0.646} & \textbf{0.455} & 0.688 & 0.523 & \textbf{0.607} & \textbf{0.399} & \textbf{0.583} & \textbf{0.359} \\
\bottomrule
\end{tabular}
\end{table}

\suppanchor{additional-metrics}
\subsection{Per-Configuration Results Across Metrics}
\label{app:additional-metrics}

\mainref{sec:setup-metrics}{5.3} reports two headline metrics and defers the
rest here.
Besides MASE and mean weighted quantile loss, the evaluator records mean and
median MSE, median MAE, MAPE, sMAPE, RMSE, NRMSE, and ND for every
configuration. We report four of these. The other four add nothing after
normalization by Seasonal Naive: relative ND equals relative MAE, relative NRMSE
equals relative RMSE, relative RMSE is the square root of relative MSE, and
median MSE equals mean MSE because our median serves as the mean forecast. Writing $y$ for an
observed future value and $\widehat y$ for the median forecast, and taking MASE
as defined in Section~\ref{app:quantile-evaluation}, the four we keep are
\begin{equation}
\mathrm{ND}=\frac{\sum|y-\widehat y|}{\sum|y|},\quad
\mathrm{MSE}=\overline{(y-\widehat y)^2},\quad
\mathrm{MAPE}=\overline{\frac{|y-\widehat y|}{|y|}},\quad
\mathrm{sMAPE}=\overline{\frac{2|y-\widehat y|}{|y|+|\widehat y|}},
\end{equation}
each averaged within a configuration and then divided by the Seasonal Naive
value for that configuration before the geometric mean over configurations
defined in the main paper is taken. Aggregation is geometric rather than
arithmetic throughout: the arithmetic mean of these ratios runs $4$ to $21\%$
higher because a few configurations dominate it. The ratios are well conditioned
even where the underlying metric is not, which matters for MAPE, whose absolute
Seasonal Naive value exceeds $10^{13}$ on configurations with near-zero targets
while its ratio stays inside $[0.10,2.66]$ across every model.

Table~\ref{tab:app-allmetrics} reports the result. \morpheusmultiscale{} leads
on MASE, ND, MSE, MAPE, and CRPS, so its advantage is not an artifact of the
seasonal scaling that MASE applies: ND normalizes by total label mass instead
and MSE not at all, and the ordering is unchanged under both. On the remaining column, sMAPE, it
ranks third. Against Falcon-2.0, the strongest prior system on MASE, it is ahead
on all six metrics, with the widest margins on MAPE ($6.6\%$) and CRPS
($6.4\%$). \morpheustwo{} is ahead of Chronos-2 on MASE, ND, MSE, and MAPE and
behind on sMAPE and CRPS; the CRPS gap reflects the narrow intervals documented
in Section~\ref{app:calibration}. Of the 97 configurations 55 are short-horizon,
so every column of Table~\ref{tab:app-allmetrics} is weighted $57\%$ toward
short requests, the regime where Section~\ref{app:gifteval-extended} places our
largest deficit.

\begin{table}[t]
\centering
\caption{All distinct GIFT-Eval metrics over 97 configurations, each normalized
by Seasonal Naive and aggregated as a geometric mean; lower is better. Relative
MAE, RMSE and NRMSE are omitted because they are determined by ND and MSE.
Bold marks the best entry in each column.}
\label{tab:app-allmetrics}
\small
\setlength{\tabcolsep}{5.0pt}
\begin{tabular}{l rrrrrr}
\toprule
Model & MASE & ND & MSE & MAPE & sMAPE & CRPS \\
\midrule
Seasonal Naive & 1.000 & 1.000 & 1.000 & 1.000 & 1.000 & 1.000 \\
Sundial-Base-128M & 0.750 & 0.739 & 0.546 & 0.777 & 0.945 & 0.559 \\
Moirai-2.0-R-Small & 0.728 & 0.705 & 0.545 & 0.693 & 0.925 & 0.516 \\
TimesFM-2.5 & 0.705 & 0.687 & 0.482 & 0.707 & 0.856 & 0.490 \\
Granite-FlowState-r1.1 & 0.701 & 0.681 & 0.493 & 0.671 & 0.833 & 0.490 \\
Chronos-2 & 0.698 & 0.679 & 0.495 & 0.666 & 0.886 & 0.485 \\
TiRex-2-Zeroshot & 0.697 & 0.673 & 0.483 & 0.641 & 0.888 & 0.478 \\
Toto-2.0-2.5B & 0.696 & 0.670 & 0.499 & 0.655 & 0.885 & 0.476 \\
Timer-S1 & 0.693 & 0.668 & 0.484 & 0.663 & 0.890 & 0.485 \\
Falcon-X & 0.687 & 0.670 & 0.476 & 0.655 & 0.886 & 0.486 \\
\textbf{\morpheustwo{}} & 0.685 & 0.673 & 0.487 & 0.653 & 0.891 & 0.519 \\
Granite-PatchTST-FM-r2 & 0.685 & 0.661 & 0.475 & 0.658 & 0.876 & 0.467 \\
Toto-2.0-2.5B-FT & 0.679 & 0.655 & 0.480 & 0.642 & 0.867 & 0.463 \\
TiRex-2-Pretrained & 0.678 & 0.659 & 0.466 & 0.651 & 0.868 & 0.467 \\
Toto-2.0-FnF & 0.676 & 0.653 & 0.463 & 0.645 & 0.867 & 0.463 \\
EXAONE-Forecast & 0.673 & 0.650 & 0.449 & 0.646 & 0.876 & 0.460 \\
TimesFM-3 & 0.667 & 0.643 & 0.444 & 0.628 & \textbf{0.796} & 0.456 \\
Falcon-2.0 & 0.666 & 0.648 & 0.462 & 0.636 & 0.861 & 0.486 \\
\textbf{\morpheusmultiscale{}} & \textbf{0.646} & \textbf{0.625} & \textbf{0.434} & \textbf{0.594} & 0.844 & \textbf{0.455} \\
\bottomrule
\end{tabular}
\end{table}

\suppanchor{domain-breakdown}
\subsection{Accuracy by Domain}
\label{app:domain-breakdown}

This section decomposes the aggregate of \mainref{sec:results}{6} by subject
matter.
Table~\ref{tab:app-allmetrics} aggregates over configurations that differ widely
in subject matter. Table~\ref{tab:app-domain} decomposes the two headline
metrics into the seven GIFT-Eval domains, normalizing within each domain by
Seasonal Naive exactly as before.

Among the systems compared, \morpheusmultiscale{} leads relative MASE on Energy,
the largest domain, and on Transport, which together hold 47 of the 97
configurations; it stays within $2$--$5\%$ of the best system on Web/CloudOps,
Econ/Fin, and Sales, and trails on Nature and Healthcare. The winning domains are the ones with many correlated variables
sampled frequently enough that several patch widths describe genuinely different
structure, the regime the mixing head targets. The losing domains
are short, sparse, and low-frequency, where the widest encoder sees little
history and the narrowest sees mostly noise. The same split holds on CRPS, where
our margin survives only on Energy and Transport while the aggregate stays the
lowest of the four. The probabilistic gap concentrates in Healthcare, Nature,
and Web/CloudOps; Chronos-2 and TimesFM-3 are both ahead of us on the first two,
and TimesFM-3 alone on the third.

\begin{table}[t]
\centering
\caption{GIFT-Eval accuracy by domain over 97 configurations. Entries are
geometric means of the per-configuration ratio to Seasonal Naive within the
domain; lower is better. Bold marks the best entry in each column within each
metric.}
\label{tab:app-domain}
\footnotesize
\setlength{\tabcolsep}{4.0pt}
\begin{tabular}{l rrrrrrr r}
\toprule
Model & Energy & Web/Cloud & Nature & Transport & Econ/Fin & Health & Sales & All \\
      & (32) & (20) & (15) & (15) & (6) & (5) & (4) & (97) \\
\midrule
\multicolumn{9}{l}{\textit{Relative MASE}} \\
TimesFM-3 & 0.798 & \textbf{0.582} & \textbf{0.687} & 0.543 & \textbf{0.739} & \textbf{0.535} & 0.682 & 0.667 \\
Chronos-2 & 0.816 & 0.611 & 0.723 & 0.603 & 0.772 & 0.551 & \textbf{0.679} & 0.698 \\
\morpheustwo{} & 0.747 & 0.614 & 0.742 & 0.557 & 0.838 & 0.655 & 0.751 & 0.685 \\
\morpheusmultiscale{} & \textbf{0.689} & 0.595 & 0.729 & \textbf{0.513} & 0.772 & 0.628 & 0.693 & \textbf{0.646} \\
\midrule
\multicolumn{9}{l}{\textit{Relative CRPS}} \\
TimesFM-3 & 0.567 & \textbf{0.418} & \textbf{0.326} & 0.402 & \textbf{0.717} & \textbf{0.400} & 0.411 & 0.456 \\
Chronos-2 & 0.585 & 0.469 & 0.340 & 0.451 & 0.735 & 0.423 & \textbf{0.406} & 0.485 \\
\morpheustwo{} & 0.570 & 0.527 & 0.398 & 0.438 & 0.825 & 0.603 & 0.482 & 0.519 \\
\morpheusmultiscale{} & \textbf{0.497} & 0.450 & 0.361 & \textbf{0.385} & 0.755 & 0.513 & 0.421 & \textbf{0.455} \\
\bottomrule
\end{tabular}
\end{table}

\suppanchor{calibration}
\subsection{Calibration of the Quantile Grid}
\label{app:calibration}

This section supports the coverage figures of \mainref{sec:limits}{8}.
Relative CRPS scores sharpness and calibration together, so it cannot say
whether the emitted intervals are the right width. We compare the empirical
frequency $\Pr(y\le\widehat q_\ell)$ against the nominal level $\ell$ at each of
the nine benchmark levels, macro-averaged over the 97 configurations
(Table~\ref{tab:app-calibration}).

For \morpheustwo{}, every level below the median over-covers and every level
above it under-covers, so the predictive distribution is too narrow rather than
displaced: its nominal $80\%$ interval covers under half of the observations.
Pooling the encoders' distributions widens the grid, as
\mainref{sec:method-multiscale}{4.6} anticipates: \morpheusmultiscale{} lifts
that interval from $49.3\%$ to $75.7\%$ and cuts the largest deviation from
nominal from $0.168$ to $0.052$, leaving its intervals only slightly narrow.
The remaining shortfall grows mildly with the requested horizon, from $76.9\%$
at short to $73.8\%$ at long, consistent with median feedback carrying no
uncertainty forward across rollout steps.

For a single encoder the defect is one of width rather than location, so the
predictive spread is the most direct route to better probabilistic accuracy,
whether by scoring a continuous ranked criterion rather than a finite pinball
sum or by tying the grid to a parameterized width. Both can be applied without
changing the architecture.

\begin{table}[t]
\centering
\caption{Calibration over 97 configurations at the nine benchmark quantile
levels. Interval columns give the empirical coverage of the central interval at
the stated nominal level; the nominal value is the target. Max.\ deviation is
the largest gap between empirical and nominal coverage over the nine levels.}
\label{tab:app-calibration}
\small
\setlength{\tabcolsep}{6.0pt}
\begin{tabular}{l rr r}
\toprule
& \multicolumn{2}{c}{Central interval coverage} & \\
\cmidrule(lr){2-3}
System & $60\%$ & $80\%$ & Max.\ dev. \\
\midrule
\morpheustwo{}        & 34.3 & 49.3 & 0.168 \\
\morpheusmultiscale{} & \textbf{53.1} & \textbf{75.7} & \textbf{0.052} \\
\bottomrule
\end{tabular}
\end{table}

\end{document}